\documentclass[]{ultrashape}

\usepackage[toc,page,header]{appendix}

\usepackage{mathrsfs}
\usepackage{adjustbox}
\usepackage{multirow}
\usepackage{multirow}
\usepackage{multicol}
\usepackage{tcolorbox}
\usepackage{changepage}
\usepackage{graphicx}
\usepackage{amssymb}
\usepackage{array}
\usepackage{bm}
\usepackage{hyperref}
\usepackage{xspace}
\usepackage{minitoc}
\usepackage{wrapfig}

\newcommand{\shortname}{UltraShape 1.0\xspace}

\title{UltraShape 1.0: High-Fidelity 3D Shape Generation via Scalable Geometric Refinement}

\author[1]{Tanghui Jia\textsuperscript{*}}
\author[2]{Dongyu Yan\textsuperscript{*}}
\author[3]{Dehao Hao\textsuperscript{*}}
\author[2]{Yang Li}
\author[3]{Kaiyi Zhang}
\author[1]{Xianyi He}
\author[2]{Lanjiong Li}
\authorbreak
\author[5]{Yuhan Wang}
\author[4]{Jinnan Chen}
\author[2]{Lutao Jiang}
\author[1]{Qishen Yin}
\author[3]{Long Quan}
\author[2]{Ying-Cong Chen}
\author[1]{Li Yuan}

\affiliation[1]{Shenzhen Graduate School, Peking University}
\affiliation[2]{The Hong Kong University of Science and Technology (Guangzhou)}
\affiliation[3]{The Hong Kong University of Science and Technology}
\affiliation[4]{National University of Singapore}
\affiliation[5]{S-Lab, Nanyang Technological University}

\contribution[*]{Equal contribution}

\abstract{
    In this report, we introduce \textbf{\shortname}, a scalable 3D diffusion framework for high-fidelity 3D geometry generation.
    The proposed approach adopts a two-stage generation pipeline: a coarse global structure is first synthesized and then refined to produce detailed, high-quality geometry.
    To support reliable 3D generation, we develop a comprehensive data processing pipeline that includes a novel watertight processing method and high-quality data filtering.
    This pipeline improves the geometric quality of publicly available 3D datasets by removing low-quality samples, filling holes, and thickening thin structures, while preserving fine-grained geometric details.
    To enable fine-grained geometry refinement, we decouple spatial localization from geometric detail synthesis in the diffusion process.
    We achieve this by performing voxel-based refinement at fixed spatial locations, where voxel queries derived from coarse geometry provide explicit positional anchors encoded via RoPE, allowing the diffusion model to focus on synthesizing local geometric details within a reduced, structured solution space.
    Our model is trained exclusively on publicly available 3D datasets, achieving strong geometric quality despite limited training resources.
    Extensive evaluations demonstrate that \shortname performs competitively with existing open-source methods in both data processing quality and geometry generation.
    All code and trained models will be released to support future research.
}

\date{\today}

\checkdata[Project Page]{\url{https://pku-yuangroup.github.io/UltraShape-1.0/} }

\begin{document}

\titlefigure{%
  \begin{center}
  \includegraphics[width=1.0\textwidth]{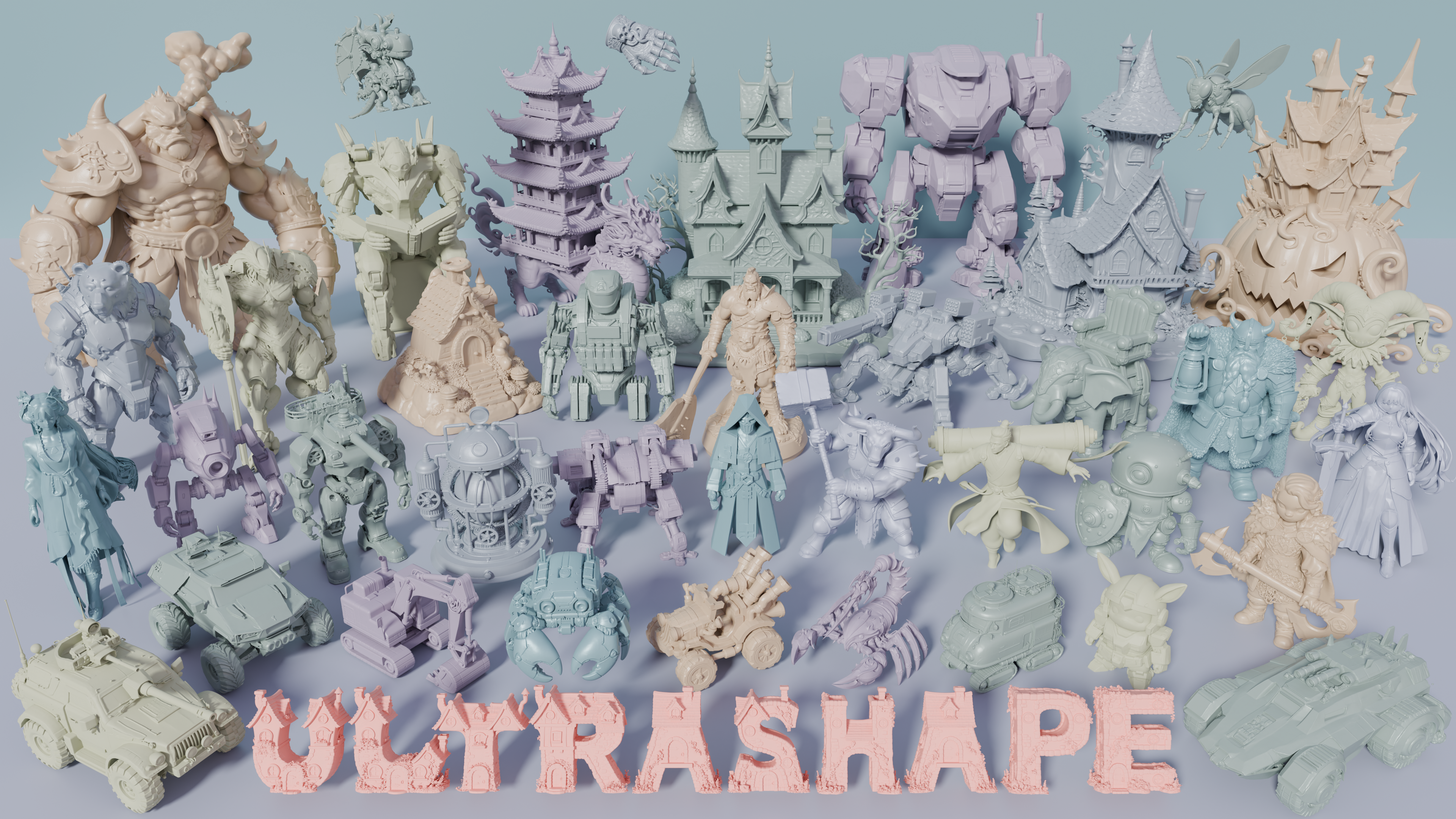}
  \captionof{figure}{High-quality 3D assets generated by \shortname. \textit{Best viewed with zoom-in.}}
  \end{center}
}

\maketitle

\section{Introduction}

3D content generation plays a fundamental role across a wide range of applications, including film and visual effects production, augmented and virtual reality, robotics, industrial design, and modern video games.
Across these domains, generation of high-fidelity 3D geometry remains a core technical requirement.
As demand for scalable, automated 3D geometry generation continues to grow, learning-based 3D generation has emerged as a key research direction in computer vision and computer graphics.
Compared to 2D content generation, 3D generation poses substantially greater challenges.
First, high-quality 3D data is significantly scarcer, often represented non-uniformly, and typically requires strong geometric properties, such as watertightness, to be directly usable in downstream tasks.
In addition, common 3D representations are inherently sparse, and both memory consumption and computational cost scale cubically with spatial resolution, severely limiting the achievable level of geometric detail and scalability.
These factors make it difficult for existing methods to produce fine-grained geometry while maintaining robustness at higher resolutions.
As a result, 3D generation techniques have not yet converged on a unified, scalable pipeline.

Existing watertight remeshing techniques for 3D generative models can be broadly categorized into UDF-based, visibility-check-based, and flood-fill-based approaches. 
UDF-based methods typically compute unsigned distance fields (UDFs) on dense voxel grids and derive pseudo-SDFs by subtracting a small offset~$\epsilon$~\cite{dora, zhang2024clay}; however, this heuristic lacks explicit sign inference, often resulting in double-layered surfaces or the erroneous removal of valid disconnected components (e.g., wheels) when filtering for the largest connected part. 
Alternatively, visibility-check-based methods employ ray casting to identify interior regions~\cite{zhang2024clay, step1x3d, craftsman}, which effectively seal cracks and eliminate spurious internal structures but remain sensitive to occlusions and prone to high-frequency geometric noise in complex regions. 
Finally, flood-fill-based strategies infer signs by expanding from exterior seeds (e.g., ManifoldPlus~\cite{huang2020manifoldplus}) to generate clean, regularized surfaces. Despite their effectiveness on closed shapes, these methods rely heavily on watertight assumptions; when applied to non-watertight or self-intersecting inputs, the fill process often leaks into the interior, yielding unintended double-layered thin shells.

Alongside earlier approaches such as Score Distillation Sampling~\cite{poole2022dreamfusion, wang2023prolificdreamer, chen2023fantasia3d} and Large Reconstruction Models~\cite{hong2023lrm, tang2024lgm, xu2024instantmesh}, diffusion transformer (DiT~\cite{peebles2023scalable})-based methods have recently become the leading paradigm in 3D generation.
They can be broadly categorized according to their underlying representations.
One major line of work adopts vector set–based representations, encoding 3D shapes as a compact set of tokens, with representative methods including 3DShape2VecSet~\cite{zhang20233dshape2vecset}, CLAY~\cite{zhang2024clay}, Hunyuan3D 2.0~\cite{zhao2025hunyuan3d}, FlashVDM~\cite{lai2025unleashing}, and TripoSG~\cite{li2025triposg}.
Vector set–based methods employ a global representation.
Although this allows objects to be expressed with relatively few tokens, it constrains their capacity to model fine-grained geometric details, often resulting in over-smoothed surfaces or missing local structures.
In contrast, another line of work focuses on sparse voxel–based representations, as exemplified by TRELLIS~\cite{xiang2025structured}, TRELLIS.2~\cite{xiang2025native}, TripoSF~\cite{he2025sparseflex}, Sparc3D~\cite{li2025sparc3d}, Hi3DGen~\cite{ye2025hi3dgen}, and Direct3D-S2~\cite{wu2025direct3d}.
These methods allocate tokens to spatially localized regions, enabling more accurate modeling of local geometry and high-frequency details.
Nevertheless, the substantially larger number of tokens introduces significant computational and memory overhead.
To mitigate this issue, sparse voxel–based approaches typically rely on two-stage generation pipelines: a sparse structure is first generated, and the structured latents are then further denoised.
By combining the two representations, Ultra3D~\cite{chen2025ultra3d} adopts a coarse-to-fine framework that refines the vector set-generated geometry using sparse voxel representations.
More recently, LATTICE~\cite{lai2025lattice} employs a voxel set formulation, providing structured queries for vector set–based models.
By introducing spatial organization into the token space, it significantly improves both scalability and geometric fidelity.

Despite their advances, existing 3D data remeshing approaches and open-sourced 3D generation methods have yet to fully address the scalability challenges inherent to high-resolution 3D geometric creation.
As a result, current approaches remain difficult to deploy directly in large-scale 3D production pipelines, limiting their practical applicability in industrial settings.
To address the aforementioned challenges of data quality and scalable geometry generation, we present \shortname, a 3D diffusion framework that jointly advances data curation and generative modeling.
On the data side, we introduce a robust framework for watertight geometry processing and a comprehensive data filtering strategy. These components respectively resolve topological ambiguities before surface extraction and ensure high-quality data curation, ultimately boosting the quality of geometric generation.
On the generation side, we adopt a two-stage coarse-to-fine strategy following LATTICE~\cite{lai2025lattice}, where a coarse global structure is first generated and then refined via voxel-conditioned diffusion.
By formulating refinement on structured voxel queries with explicit spatial encoding, the second stage decouples spatial localization from geometric detail synthesis, enabling stable training and fine-grained geometry generation at scale.

\section{Method}

\subsection{Data Curation Pipeline}

\paragraph{Data Watertightening.}
Watertight remeshing plays a critical role in geometry generative modeling.
On the one hand, watertightness guarantees a globally well-defined interior–exterior partition, making volumetric representations such as signed distance fields (SDFs) semantically meaningful. On the other hand, watertight remeshing serves not merely as a geometric repair operation. Still, as a form of geometric standardization, it removes statistically noisy and semantically irrelevant structures, such as spurious internal components or modeling artifacts that are weakly correlated with the object’s outer surface, yielding a cleaner and more learnable geometric signal.

Inspired by watershed algorithms, we develop a novel voxel-based reconstruction approach for watertight geometry processing.
The method operates in a sparse volumetric domain, where topological ambiguities can be resolved robustly before surface extraction. Its key features are summarized as follows:
\begin{enumerate}
    \item \textbf{Scalable CUDA-Parallel Sparse Voxel Infrastructure.}  
    CUDA-parallel sparse data structures and algorithms, enabling scalable voxel reconstruction at resolutions up to $2048^3$.

    \item \textbf{Robust Automatic Hole Closing.}  
    Automatic hole closing, which robustly seals gaps and cracks commonly found in real-world meshes.

    \item \textbf{Open-Surface Identification and Volumetric Thickening.}  
    Automatic volumetric thickening of open surfaces is enabled by our open-surface identification method, which detects and resolves the zero-volume issue in open meshes before signed distance field reconstruction.
\end{enumerate}

\paragraph{Data Filtering.}
Our initial data pool comes from Objaverse~\cite {objaverse}, which contains about 800K 3D models across diverse categories and styles. However, after coarse inspection, we noticed unsatisfactory data quality and identified three major issues that would hinder the training: \textbf{Low-Quality Geometry}. A significant portion is meaningless, consists of simple primitives, and includes objects with inconsistent geometry and texture, as well as poor modeling quality. Additionally, many scanned models exhibit fragmented meshes and incomplete topologies, which are prohibitive for watertightness and result in excessively thin structures. \textbf{Inconsistent Poses}. 3D assets within the same categories show erratic orientations, especially for humanoid models. We found that such misalignment impedes DiT from learning coherent shape priors, thereby inducing instability during training. \textbf{Intricate Internal Structures}. Many hand-crafted models are manually assembled from disconnected components, leaving tiny seams and self-intersecting faces. These meshes are prone to degenerating into hollow thin shells or fragments during watertighting. Driven by these observations, we tailored a data filtering pipeline to curate high-quality 3D models for training, which consists of the following steps:
\begin{enumerate}
    \item \textbf{VLM-based Filtering}. We rendered multiple views of each 3D model with depth and normal maps, and employed a vision-language model (VLM) to filter out simple primitives, ambient ground planes, and noisy scanned scenes. 
    \item \textbf{Pose Normalization}. We also used VLM to remove misposed models. Inspired by ~\cite{jin2025oneshot}, we also trained a pose canonicalization network to detect mis-posed models and normalize them into consistent orientations.
    \item \textbf{Geometry Filtering}. For models with complex internal structures, we first calculated the ratio of interior to exterior points near the watertight surfaces to identify whole thin-shells. Furthermore, we observed that the variational autoencoder (VAE) reconstruction of these models often results in severe fragmentation. Therefore, we employed a pretrained VAE to filter out models that exhibit numerous disconnected components after reconstruction. 
\end{enumerate}

Following the automated filtering process, we conducted a manual inspection phase to ensure the integrity of the final data further. In total, our curation pipeline refined the initial 800K models down to approximately 330K valid samples, of which 120K were identified as high-quality. This curated dataset proved sufficient for training our proposed two-stage generation framework.

\subsection{Geometry Generation}

\begin{figure}[t]
    \centering
    \includegraphics[width=1.0\textwidth]{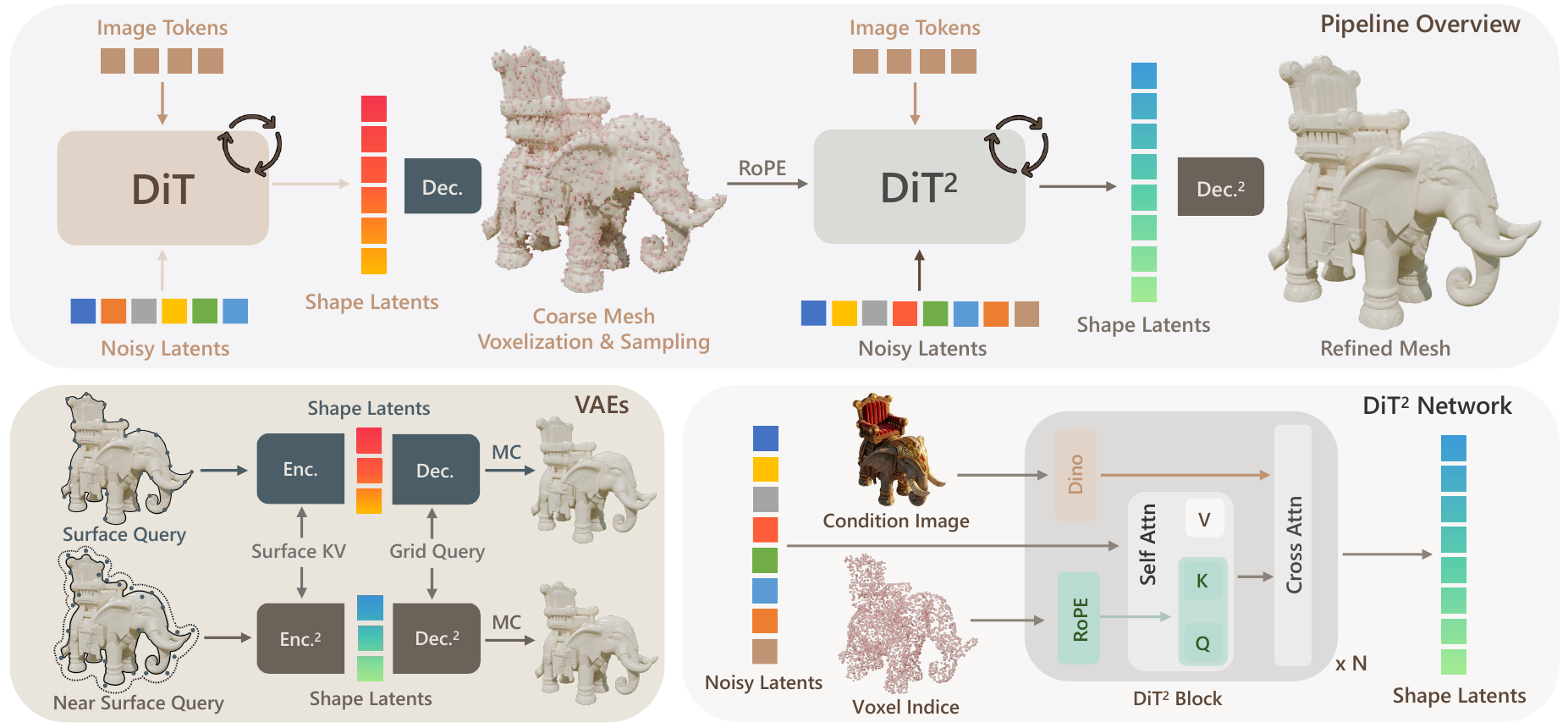}
    \caption{\textbf{Overview of \shortname pipeline}, where Enc. and Dec. represent the encoder and decoder of our VAE. The superscript “2” on the upper right corner denotes the Stage-2 model. MC means  marching cube.}
    \label{fig:pipeline}
\end{figure}

Following LATTICE~\cite{lai2025lattice}, our geometry generation framework adopts a two-stage coarse-to-fine design to balance global structural coherence and fine-grained geometric detail.
We refer readers to~\cite{lai2025lattice} for a comprehensive discussion of the overall formulation, and provide a brief description here for completeness.
Specifically, we first generate a coarse representation that captures the overall shape of the object, and then refine it using voxel-based queries to synthesize detailed, high-quality geometry.
In addition, we explore the potential of the proposed representation for training-free 3D stylization.

\paragraph{Coarse Structure Generation.}
The primary objective of the first stage is to provide reliable and informative voxel queries for the subsequent refinement stage.
Rather than targeting fine-grained geometric details, the first stage is designed to capture the object's overall structure, providing a coarse yet semantically meaningful geometric representation that can be effectively refined in the second stage.
To serve this purpose, the generated geometry must exhibit strong global shape awareness and robust generalization across diverse object categories.
This coarse structural prior enables the second-stage model to focus on geometry refinement and detail synthesis without being burdened by global structural ambiguities.
Based on these considerations, we adopt a DiT-based 3D generation model operating on a vector set representation as the first-stage generator, which provides a compact and expressive encoding of global object geometry suitable for downstream refinement.

\begin{figure}[t]
    \centering
    \includegraphics[width=1.0\textwidth]{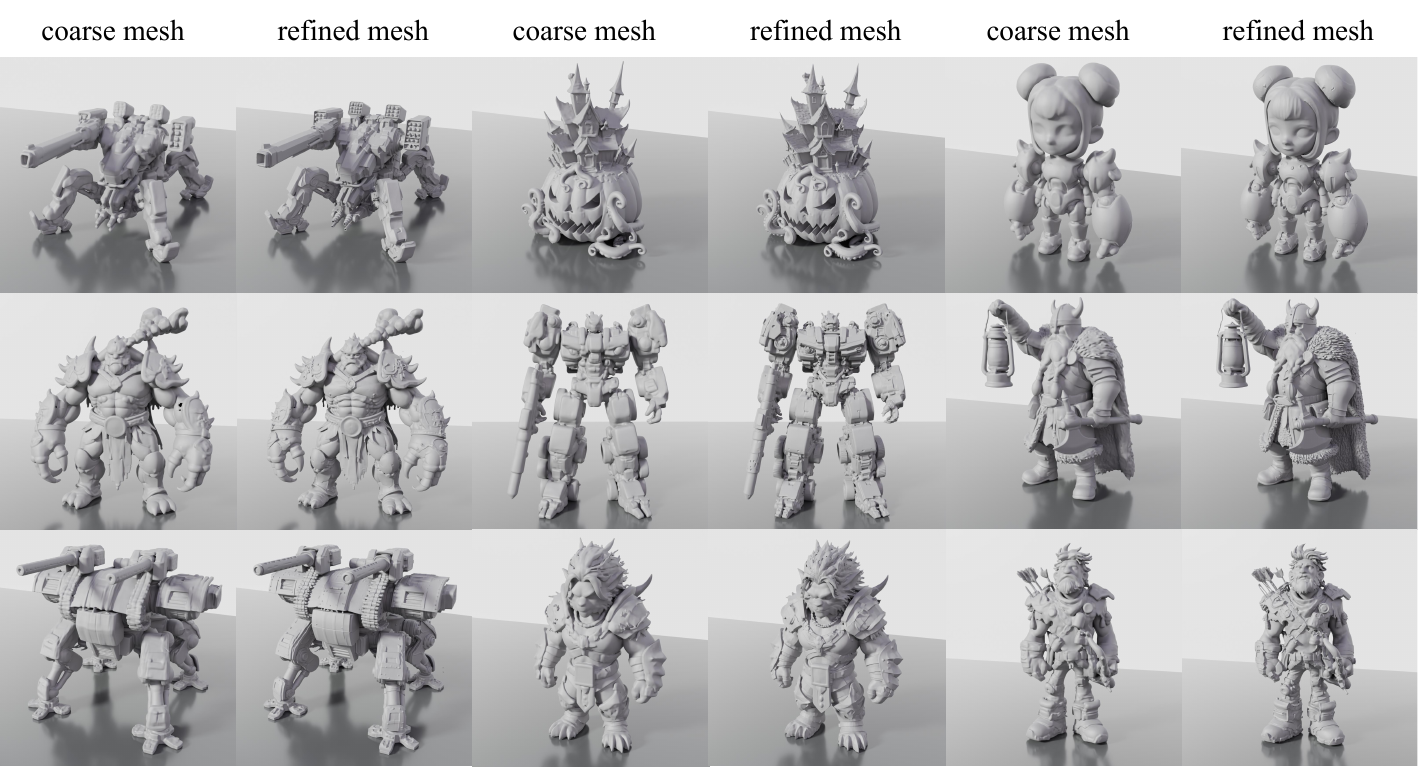}
    \caption{Comparison of our refined mesh against the coarse mesh generated from the first stage.
    \textit{Best viewed with zoom-in.}}
    \label{fig:refine}
\end{figure}

\paragraph{Geometry Refinement.}
LATTICE~\cite{lai2025lattice} has analyzed why existing vector set–based 3D diffusion models often struggle to generate fine-grained geometric details.
In particular, two closely related factors have been identified as key contributors to this limitation:
\begin{enumerate}
    \item Vector set–based methods typically rely on VAEs with surface queries.
    As a result, latent vectors correspond to spatial locations distributed across the entire volume, leading to an inherently large and unstructured solution space for the diffusion model during generation.
    Such a large, unstructured latent space makes the diffusion process more difficult to converge, especially when modeling high-frequency geometric details.
    \item In vector set–based approaches, the diffusion model is required to simultaneously generate both positional information and local geometric features within the latent representation.
    In contrast, sparse voxel–based methods condition on fixed spatial locations and primarily focus on synthesizing local geometry.
    This coupling of global positioning and local shape synthesis further increases the complexity of the diffusion task in vector set–based settings.
\end{enumerate}

With the analysis described above, the refinement stage is formulated to explicitly decouple spatial localization from geometric detail synthesis.
This is achieved by performing diffusion-based refinement on voxel queries defined over a fixed-resolution grid, which constrains generation to a structured and discretized spatial domain.
As a result, the diffusion process operates in a substantially reduced and more stable solution space.
In this coarse-to-fine formulation, coarse geometry provides explicit spatial anchors for refinement.
Voxel queries derived from the coarse shape define fixed spatial locations, whose coordinates are encoded using rotary positional embeddings (RoPE)~\cite{su2024roformer}.
With spatial localization explicitly specified, the diffusion model focuses on synthesizing local geometric details rather than jointly modeling global positioning and shape, leading to improved convergence and finer geometric refinement.
Compared to surface-based queries, voxel-point queries allow further geometric refinement without strictly adhering to the coarse input.

\begin{wrapfigure}{r}{0.75\textwidth}
    \centering
    \vspace{-1mm}
    \includegraphics[width=0.75\textwidth]{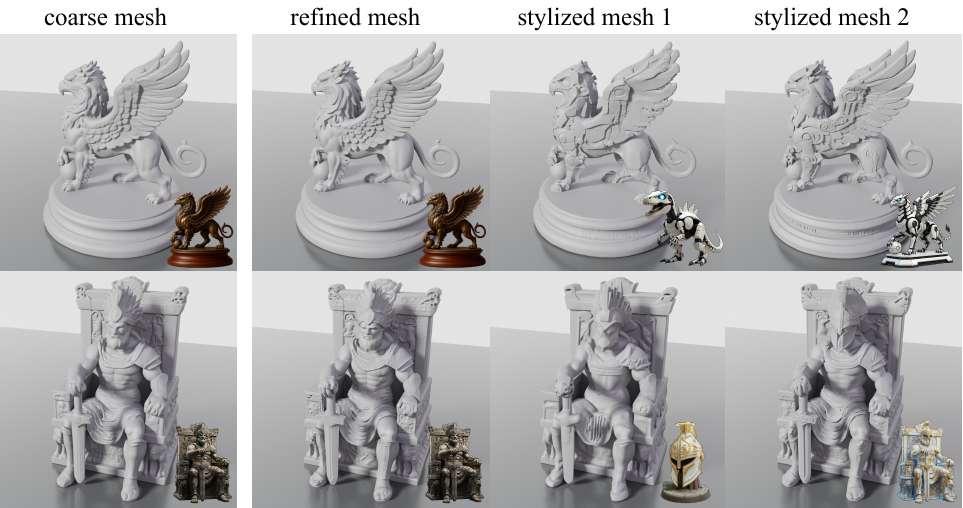}
    \caption{Stylized results generated using our training-free stylization method. The refined and stylized meshes in the last three columns are generated by refining the coarse mesh in the first column using image conditions on the bottom-right corner. \textit{Best viewed with zoom-in.}}
    \vspace{-1mm}
    \label{fig:stylization}
\end{wrapfigure}

To support voxel-based refinement, the shape VAE is extended to decode geometry at off-surface locations.
During training, surface queries are augmented with bounded spatial perturbations, enabling the decoder to predict valid volumetric geometry.
At inference time, voxel queries sampled from the coarse geometry are aligned with latent tokens and refined through a diffusion process.
The denoised latent representation is decoded into an SDF field on a regular grid, from which the final surface is extracted using marching cubes.
The refinement stage employs a DiT architecture with self-attention over latent tokens, and the results are shown in Fig. \ref{fig:refine}.
Spatial information is injected via RoPE at each layer, while image conditioning is incorporated through cross-attention using DINOv2~\cite{oquab2023dinov2} features.
An image token masking strategy is applied to suppress irrelevant background information, ensuring robust and semantically aligned geometry refinement.

\paragraph{Training-Free Stylization.}
We further discovered the potential of training-free stylization using voxel-conditioned latent.
Specifically, by conditioning on different images in the two stages, we can generate 3D geometry that follows the coarse shape from the image used in the first stage and finer stylized details from the image used in the second stage, as shown in Fig. \ref{fig:stylization}.
The results show that, despite differences in the conditions used in the two stages, the coarse voxel representation enables the second stage to perform fine-detail sculpting without introducing conflicts.

\begin{figure}[t]
    \centering
    \includegraphics[width=1.0\textwidth]{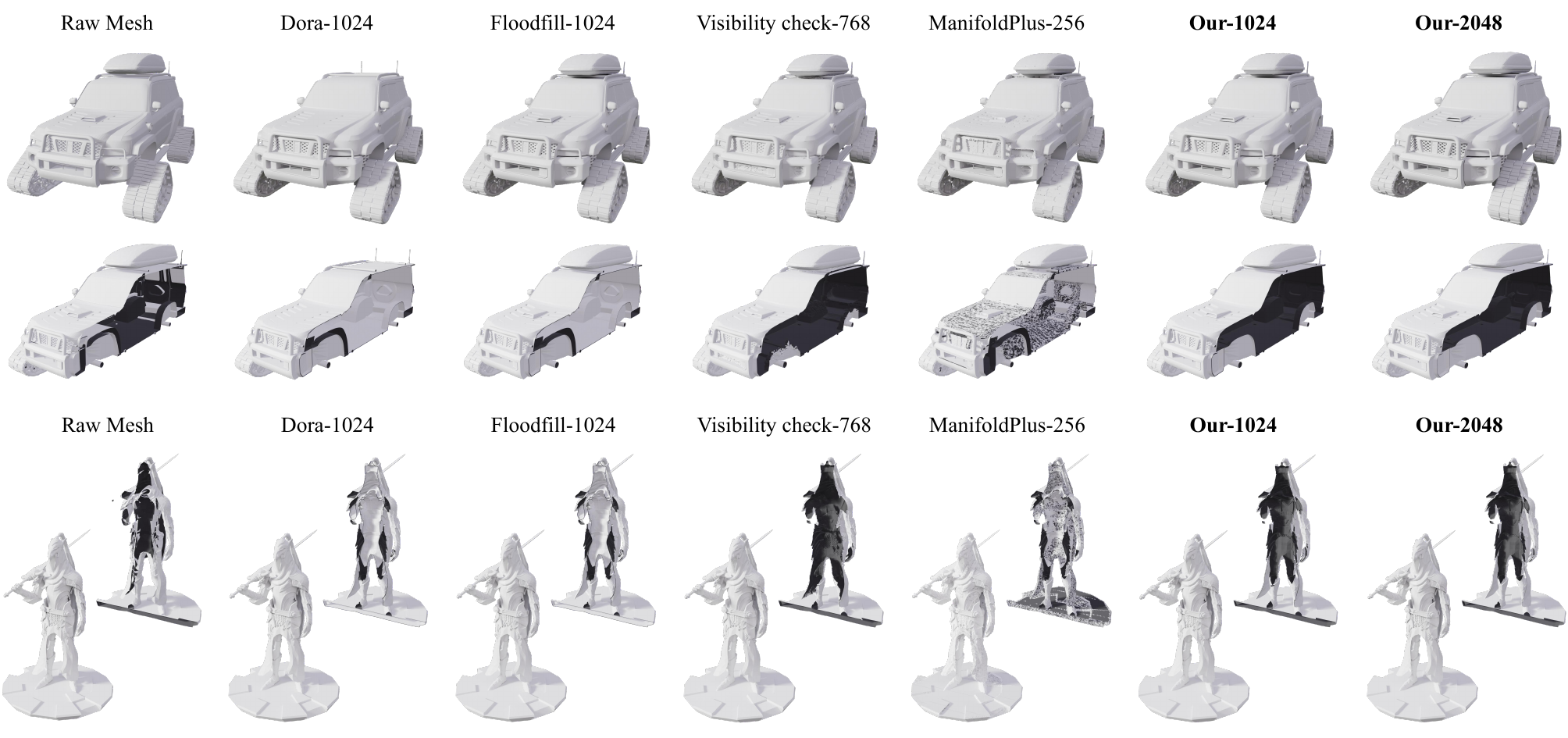}
    \caption{Comparison of our watertightening method against other approaches. \textit{Best viewed with zoom-in.}}
    \label{fig:watertight}
\end{figure}

\section{Experiments}

\subsection{Implementation Details}

\paragraph{Data Preparation.}
We first sample points for both the VAE input and supervision.
Specifically, we sample surface points uniformly across the mesh, placing additional emphasis on sharp regions with high curvature to better preserve geometric details.
These surface points are used as inputs to the VAE encoder.
For supervision, we sample points in the vicinity of the surface, including uniformly sampled near-surface points and curvature-aware sharp points.
We also randomly sample supervision points in free space.
SDF values are computed for all supervision points and used to define the reconstruction loss.
For each object, we sample approximately 600K surface points for VAE input and 1M points for supervision.
For image rendering, we use the Cycles renderer in Blender with orthographic projection.
All images are rendered at a resolution of $1024^2$.
For each object, we render 16 images: eight from near-frontal viewpoints and eight from randomly sampled orientations to increase viewpoint diversity.
To further enhance visual variability and robustness, we randomly select environment maps for lighting augmentation during rendering.

\paragraph{Detailed Settings}
For coarse structure generation, we directly adopt Hunyuan3D-2.1~\cite{hunyuan3d2025hunyuan3d} as the first-stage model, as it demonstrates strong generalization across diverse object categories among existing open-source approaches.
The VAE used in the refinement stage is initialized from the Hunyuan3D-2.1 VAE and fine-tuned for 55K steps with a uniform query perturbation sampled from $[-1/128, 1/128]$.
Training is conducted progressively: we first fine-tune for 40K steps with 4096 tokens, followed by 15K steps with 8192 tokens, thereby improving training stability and enabling generalization to higher token counts.
The diffusion transformer (DiT) for geometry refinement is also initialized from Hunyuan3D-2.1 and fine-tuned on our dataset.
We use a voxel resolution of 128 for both training and inference, and adopt a progressive multi-stage strategy that jointly increases token count and image resolution: (1) 4096 tokens at 518 resolution for 10K steps; (2) 8192 tokens at 1022 resolution for 15K steps; and (3) 10240 tokens at 1022 resolution for 60K steps.
All experiments are conducted on 8 NVIDIA H20 GPUs with a batch size of 32, using 120K filtered samples from Objaverse.
During inference, we use 32768 tokens and an image resolution of 1022 with token masking unless otherwise specified.
We further observe that input RGBA image quality is critical: inaccurate foreground segmentation or residual background and shadow artifacts can noticeably degrade geometry quality, highlighting the importance of robust image pre-processing for image-conditioned 3D generation.

\subsection{Data Watertightening}

As shown in Figure~\ref{fig:watertight}, existing watertight reconstruction methods struggle to achieve robust hole filling and high-quality geometry simultaneously. Flood-fill-based methods and Dora-style approaches often fail to close large open surfaces, leading to missing volumes or incomplete shapes. Visibility-based methods can seal open regions but often rely on heuristic inside–outside tests, which introduce noisy signed distance fields and irregular geometry, particularly in thin structures and self-occluded regions.

In contrast, our method consistently produces watertight reconstructions with significantly improved geometric quality. It successfully closes open surfaces without introducing the noisy artifacts commonly observed in visibility-based approaches. Across all examples, our results exhibit cleaner surfaces, better preservation of fine details, and more stable geometry, demonstrating the effectiveness of our approach for robust watertight geometry processing.

\begin{figure}[htbp]
    \centering
    \includegraphics[width=1.0\textwidth]{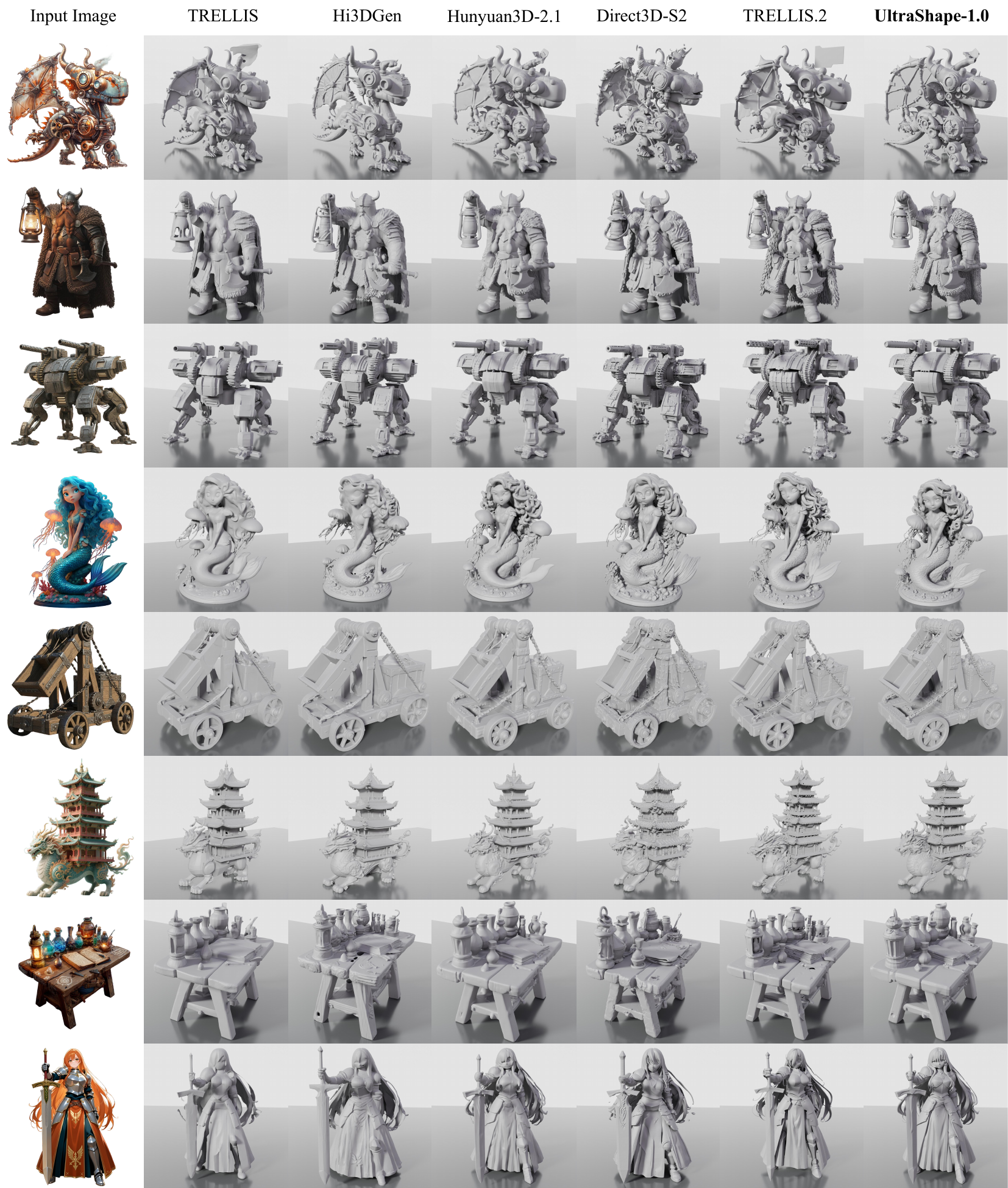}
    \caption{Comparison of \shortname against SOTA open-sourced methods. \textit{Best viewed with zoom-in.}}
    \label{fig:dit_comp_open1}
\end{figure}

\begin{figure}[htbp]
    \centering
    \includegraphics[width=1.0\textwidth]{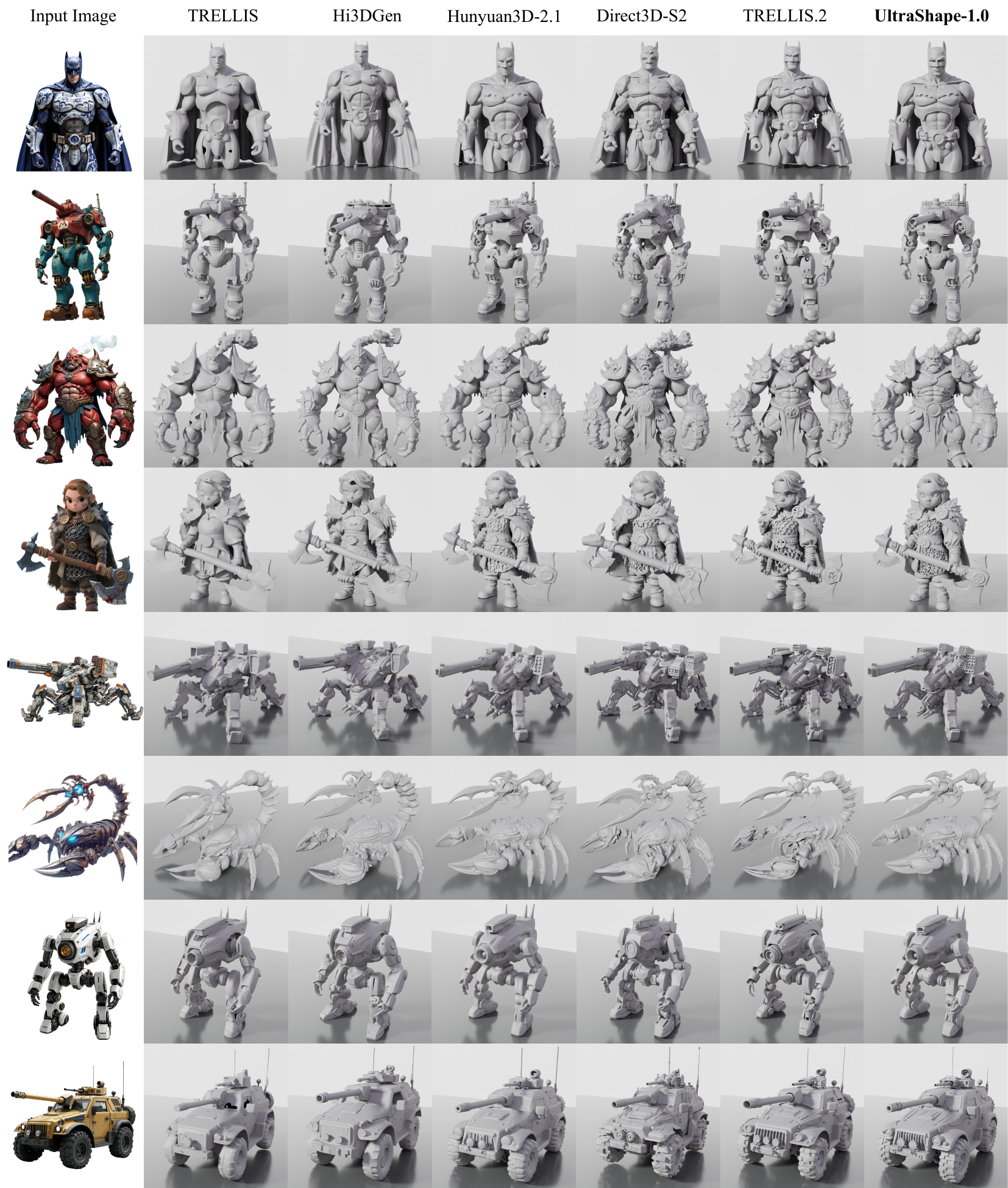}
    \caption{Comparison of \shortname against SOTA open-sourced methods. \textit{Best viewed with zoom-in.}}
    \label{fig:dit_comp_open2}
\end{figure}

\begin{figure}[htbp]
    \centering
    \includegraphics[width=1.0\textwidth]{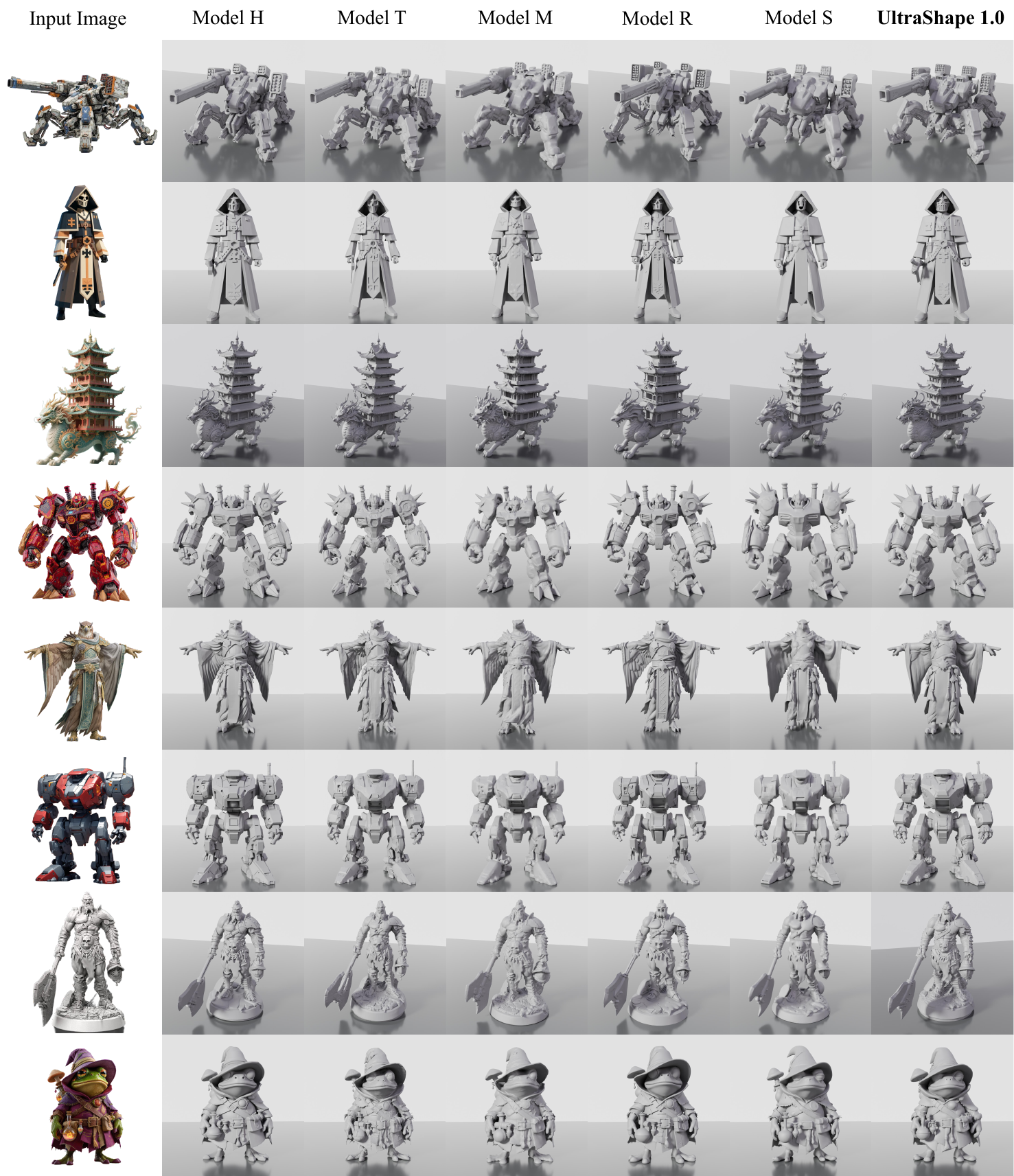}
    \caption{Comparison of \shortname against commercial methods. \textit{Best viewed with zoom-in.}}
    \label{fig:dit_comp_commercial}
\end{figure}

\clearpage

\subsection{Shape Generation}

\begin{wrapfigure}{r}{0.66\textwidth}
    \centering
    \includegraphics[width=0.66\textwidth]{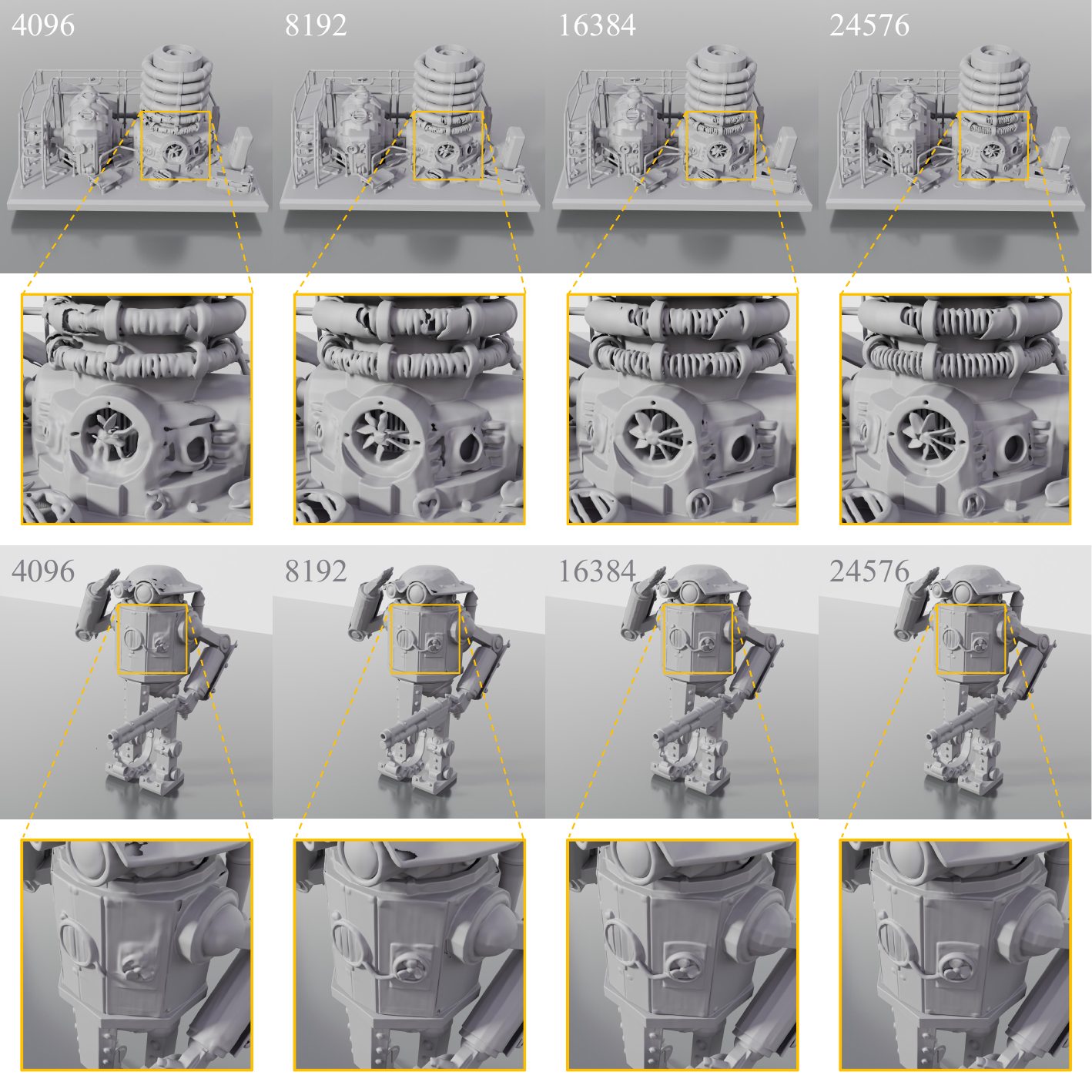}
    \caption{Comparison of VAE inference results when extrapolating the number of tokens during reconstruction. \textit{Best viewed with zoom-in.}}
    \label{fig:vae_extrapolation}
\end{wrapfigure}

We first evaluate the test-time scalability of our model, showing results in  Fig. \ref{fig:vae_extrapolation}.
As the number of latent tokens increases, reconstruction quality consistently improves, demonstrating strong potential to reconstruct high-fidelity geometries.

We then evaluate the generation performance of our model and compare it with representative open-source state-of-the-art approaches.
Qualitative results are shown in Figs. \ref{fig:dit_comp_open1} and \ref{fig:dit_comp_open2}.
As shown, our model produces high-quality 3D geometry with rich details and sharp structural features, while maintaining strong consistency with the input condition image.
Compared to existing baselines, our results exhibit superior geometric fidelity and improved alignment between geometry and visual appearance.

We further compare our model with commercial 3D generation models, with qualitative results presented in Fig. \ref{fig:dit_comp_commercial}.
Despite being trained on only publicly available data and limited computational resources, our model achieves generation quality comparable to that of commercial systems in both geometric detail and image–geometry consistency.
These results demonstrate the strong effectiveness and competitiveness of our approach.

\begin{figure}[htbp]
    \centering
    \includegraphics[width=1.0\textwidth]{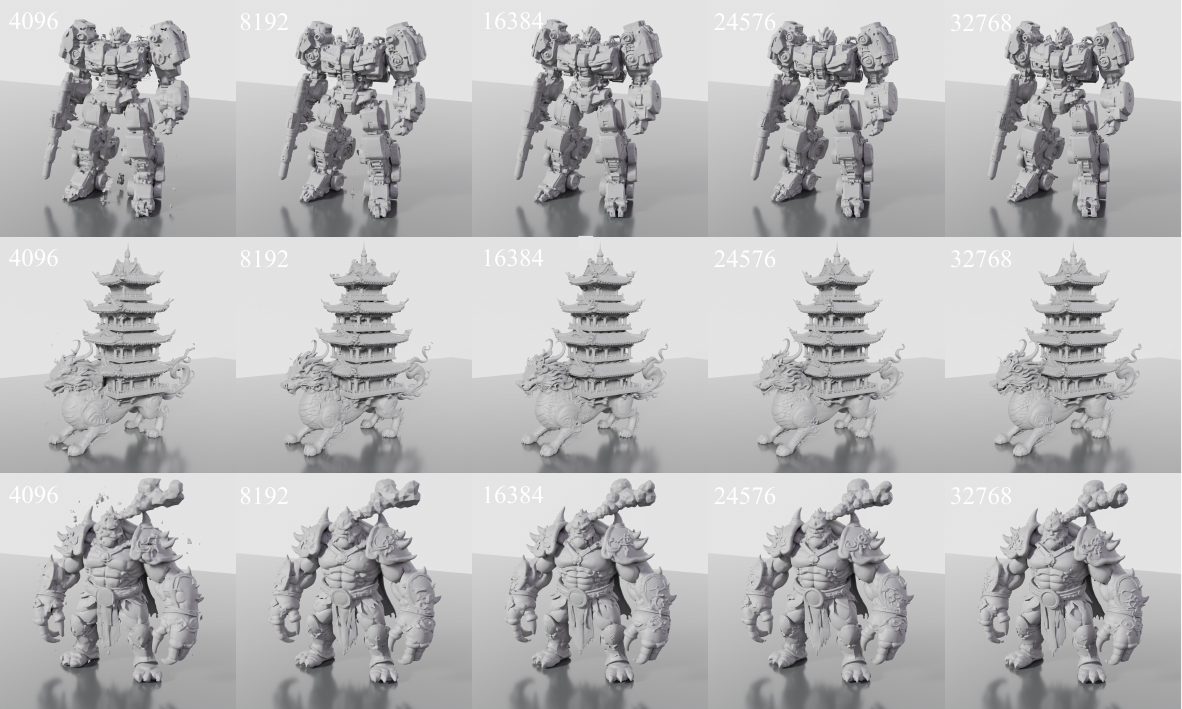}
    \caption{Comparison of DiT inference results when extrapolating the number of tokens during generation. \textit{Best viewed with zoom-in.}}
    \label{fig:dit_extrapolation}
\end{figure}

Finally, similar to the VAE reconstruction experiments, we also investigate the test-time scaling behavior of the DiT used in geometry generation.
Fig. \ref{fig:dit_extrapolation} illustrates the effect of increasing the number of shape tokens and image tokens during inference.
Although the model is trained with a relatively small token budget, it generalizes well to significantly larger token counts at test time, producing substantially improved geometric details and higher-quality surfaces.
This result highlights the favorable test-time scalability of the generation framework.

\section{Conclusion}

In this report, we presented \shortname, a scalable and high-fidelity 3D diffusion framework for geometric asset generation.
To address the challenges of data quality, geometric scalability, and fine-detail synthesis in existing 3D generation methods, we proposed a two-stage geometry-generation pipeline that decouples global-structure modeling from local-detail refinement.
Our approach combines a robust data-processing pipeline with a coarse-to-fine geometry-generation strategy.
Through a novel, watertight data-processing method and high-quality data filtering, we improve the reliability of the training data while preserving geometric details.
For geometry generation, we leverage a diffusion-based vector set representation to capture global structure in the first stage, and introduce a voxel-based refinement stage that significantly enhances geometric detail and scalability.
By conditioning the refinement process on spatially localized voxel queries and injecting coarse geometry information via positional encoding, our method enables stable training and high-quality geometry synthesis at higher resolutions.
Experimental results demonstrate that \shortname achieves strong performance compared to existing open-source 3D generation methods in both data processing quality and geometry generation.
Notably, our approach achieves high-quality geometric generation with low-cost training resources.
Our work provides a meaningful step toward more scalable and practical 3D generation systems, and offers valuable insights for future research on large-scale 3D generative models and their real-world deployment.

\clearpage

\bibliographystyle{plainnat}
\bibliography{main}

\begin{thebibliography}{29}
\providecommand{\natexlab}[1]{#1}
\providecommand{\url}[1]{\texttt{#1}}
\expandafter\ifx\csname urlstyle\endcsname\relax
  \providecommand{\doi}[1]{doi: #1}\else
  \providecommand{\doi}{doi: \begingroup \urlstyle{rm}\Url}\fi

\bibitem[Chen et~al.(2023)Chen, Chen, Jiao, and Jia]{chen2023fantasia3d}
Rui Chen, Yongwei Chen, Ningxin Jiao, and Kui Jia.
\newblock Fantasia3d: Disentangling geometry and appearance for high-quality text-to-3d content creation.
\newblock In \emph{Proceedings of the IEEE/CVF international conference on computer vision}, pages 22246--22256, 2023.

\bibitem[Chen et~al.(2025{\natexlab{a}})Chen, Zhang, Liang, Luo, Li, Liu, Li, Long, Feng, and Tan]{dora}
Rui Chen, Jianfeng Zhang, Yixun Liang, Guan Luo, Weiyu Li, Jiarui Liu, Xiu Li, Xiaoxiao Long, Jiashi Feng, and Ping Tan.
\newblock Dora: Sampling and benchmarking for 3d shape variational auto-encoders.
\newblock In \emph{Proceedings of the Computer Vision and Pattern Recognition Conference (CVPR)}, pages 16251--16261, June 2025{\natexlab{a}}.

\bibitem[Chen et~al.(2025{\natexlab{b}})Chen, Li, Wang, Zhang, Li, Zhang, and Lin]{chen2025ultra3d}
Yiwen Chen, Zhihao Li, Yikai Wang, Hu~Zhang, Qin Li, Chi Zhang, and Guosheng Lin.
\newblock Ultra3d: Efficient and high-fidelity 3d generation with part attention.
\newblock \emph{arXiv preprint arXiv:2507.17745}, 2025{\natexlab{b}}.

\bibitem[Deitke et~al.(2022)Deitke, Schwenk, Salvador, Weihs, Michel, VanderBilt, Schmidt, Ehsani, Kembhavi, and Farhadi]{objaverse}
Matt Deitke, Dustin Schwenk, Jordi Salvador, Luca Weihs, Oscar Michel, Eli VanderBilt, Ludwig Schmidt, Kiana Ehsani, Aniruddha Kembhavi, and Ali Farhadi.
\newblock Objaverse: A universe of annotated 3d objects.
\newblock \emph{arXiv preprint arXiv:2212.08051}, 2022.

\bibitem[He et~al.(2025)He, Zou, Chen, Guo, Liang, Yuan, Ouyang, Cao, and Li]{he2025sparseflex}
Xianglong He, Zi-Xin Zou, Chia-Hao Chen, Yuan-Chen Guo, Ding Liang, Chun Yuan, Wanli Ouyang, Yan-Pei Cao, and Yangguang Li.
\newblock Sparseflex: High-resolution and arbitrary-topology 3d shape modeling.
\newblock \emph{arXiv preprint arXiv:2503.21732}, 2025.

\bibitem[Hong et~al.(2023)Hong, Zhang, Gu, Bi, Zhou, Liu, Liu, Sunkavalli, Bui, and Tan]{hong2023lrm}
Yicong Hong, Kai Zhang, Jiuxiang Gu, Sai Bi, Yang Zhou, Difan Liu, Feng Liu, Kalyan Sunkavalli, Trung Bui, and Hao Tan.
\newblock Lrm: Large reconstruction model for single image to 3d.
\newblock \emph{arXiv preprint arXiv:2311.04400}, 2023.

\bibitem[Huang et~al.(2020)Huang, Zhou, and Guibas]{huang2020manifoldplus}
Jingwei Huang, Yichao Zhou, and Leonidas Guibas.
\newblock Manifoldplus: A robust and scalable watertight manifold surface generation method for triangle soups.
\newblock \emph{arXiv preprint arXiv:2005.11621}, 2020.

\bibitem[Hunyuan3D et~al.(2025)Hunyuan3D, Yang, Yang, Feng, Huang, Zhang, He, Luo, Liu, Zhao, et~al.]{hunyuan3d2025hunyuan3d}
Team Hunyuan3D, Shuhui Yang, Mingxin Yang, Yifei Feng, Xin Huang, Sheng Zhang, Zebin He, Di~Luo, Haolin Liu, Yunfei Zhao, et~al.
\newblock Hunyuan3d 2.1: From images to high-fidelity 3d assets with production-ready pbr material.
\newblock \emph{arXiv preprint arXiv:2506.15442}, 2025.

\bibitem[Jin et~al.(2025)Jin, Wang, Chen, Dai, Gao, Qin, and Chen]{jin2025oneshot}
Li~Jin, Yujie Wang, Wenzheng Chen, Qiyu Dai, Qingzhe Gao, Xueying Qin, and Baoquan Chen.
\newblock One-shot 3d object canonicalization based on geometric and semantic consistency.
\newblock In \emph{2025 IEEE/CVF Conference on Computer Vision and Pattern Recognition (CVPR)}, pages 16850--16859, 2025.
\newblock \doi{10.1109/CVPR52734.2025.01570}.

\bibitem[Lai et~al.(2025{\natexlab{a}})Lai, Zhao, Zhao, Liu, Lin, Huang, Guo, and Yue]{lai2025lattice}
Zeqiang Lai, Yunfei Zhao, Zibo Zhao, Haolin Liu, Qingxiang Lin, Jingwei Huang, Chunchao Guo, and Xiangyu Yue.
\newblock Lattice: Democratize high-fidelity 3d generation at scale.
\newblock \emph{arXiv preprint arXiv:2512.03052}, 2025{\natexlab{a}}.

\bibitem[Lai et~al.(2025{\natexlab{b}})Lai, Zhao, Zhao, Liu, Wang, Shi, Yang, Lin, Huang, Liu, et~al.]{lai2025unleashing}
Zeqiang Lai, Yunfei Zhao, Zibo Zhao, Haolin Liu, Fuyun Wang, Huiwen Shi, Xianghui Yang, Qingxiang Lin, Jingwei Huang, Yuhong Liu, et~al.
\newblock Unleashing vecset diffusion model for fast shape generation.
\newblock \emph{arXiv preprint arXiv:2503.16302}, 2025{\natexlab{b}}.

\bibitem[Li et~al.(2024)Li, Liu, Chen, Liang, Chen, Tan, and Long]{craftsman}
Weiyu Li, Jiarui Liu, Rui Chen, Yixun Liang, Xuelin Chen, Ping Tan, and Xiaoxiao Long.
\newblock Craftsman: High-fidelity mesh generation with 3d native generation and interactive geometry refiner.
\newblock \emph{arXiv preprint arXiv:2405.14979}, 2024.

\bibitem[Li et~al.(2025{\natexlab{a}})Li, Zhang, Sun, Qi, Li, Cheng, Cai, Wu, Liu, Wang, Chen, Tian, Pan, Li, Yu, Zhang, Jiang, and Tan]{step1x3d}
Weiyu Li, Xuanyang Zhang, Zheng Sun, Di~Qi, Hao Li, Wei Cheng, Weiwei Cai, Shihao Wu, Jiarui Liu, Zihao Wang, Xiao Chen, Feipeng Tian, Jianxiong Pan, Zeming Li, Gang Yu, Xiangyu Zhang, Daxin Jiang, and Ping Tan.
\newblock Step1x-3d: Towards high-fidelity and controllable generation of textured 3d assets.
\newblock \emph{arXiv preprint arxiv:2505.07747}, 2025{\natexlab{a}}.

\bibitem[Li et~al.(2025{\natexlab{b}})Li, Zou, Liu, Wang, Liang, Yu, Liu, Guo, Liang, Ouyang, et~al.]{li2025triposg}
Yangguang Li, Zi-Xin Zou, Zexiang Liu, Dehu Wang, Yuan Liang, Zhipeng Yu, Xingchao Liu, Yuan-Chen Guo, Ding Liang, Wanli Ouyang, et~al.
\newblock Triposg: High-fidelity 3d shape synthesis using large-scale rectified flow models.
\newblock \emph{arXiv preprint arXiv:2502.06608}, 2025{\natexlab{b}}.

\bibitem[Li et~al.(2025{\natexlab{c}})Li, Wang, Zheng, Luo, and Wen]{li2025sparc3d}
Zhihao Li, Yufei Wang, Heliang Zheng, Yihao Luo, and Bihan Wen.
\newblock Sparc3d: Sparse representation and construction for high-resolution 3d shapes modeling.
\newblock \emph{arXiv preprint arXiv:2505.14521}, 2025{\natexlab{c}}.

\bibitem[Oquab et~al.(2023)Oquab, Darcet, Moutakanni, Vo, Szafraniec, Khalidov, Fernandez, Haziza, Massa, El-Nouby, et~al.]{oquab2023dinov2}
Maxime Oquab, Timoth{\'e}e Darcet, Th{\'e}o Moutakanni, Huy Vo, Marc Szafraniec, Vasil Khalidov, Pierre Fernandez, Daniel Haziza, Francisco Massa, Alaaeldin El-Nouby, et~al.
\newblock Dinov2: Learning robust visual features without supervision.
\newblock \emph{arXiv preprint arXiv:2304.07193}, 2023.

\bibitem[Peebles and Xie(2023)]{peebles2023scalable}
William Peebles and Saining Xie.
\newblock Scalable diffusion models with transformers.
\newblock In \emph{Proceedings of the IEEE/CVF international conference on computer vision}, pages 4195--4205, 2023.

\bibitem[Poole et~al.(2022)Poole, Jain, Barron, and Mildenhall]{poole2022dreamfusion}
Ben Poole, Ajay Jain, Jonathan~T Barron, and Ben Mildenhall.
\newblock Dreamfusion: Text-to-3d using 2d diffusion.
\newblock \emph{arXiv preprint arXiv:2209.14988}, 2022.

\bibitem[Su et~al.(2024)Su, Ahmed, Lu, Pan, Bo, and Liu]{su2024roformer}
Jianlin Su, Murtadha Ahmed, Yu~Lu, Shengfeng Pan, Wen Bo, and Yunfeng Liu.
\newblock Roformer: Enhanced transformer with rotary position embedding.
\newblock \emph{Neurocomputing}, 568:\penalty0 127063, 2024.

\bibitem[Tang et~al.(2024)Tang, Chen, Chen, Wang, Zeng, and Liu]{tang2024lgm}
Jiaxiang Tang, Zhaoxi Chen, Xiaokang Chen, Tengfei Wang, Gang Zeng, and Ziwei Liu.
\newblock Lgm: Large multi-view gaussian model for high-resolution 3d content creation.
\newblock \emph{arXiv preprint arXiv:2402.05054}, 2024.

\bibitem[Wang et~al.(2023)Wang, Lu, Wang, Bao, Li, Su, and Zhu]{wang2023prolificdreamer}
Zhengyi Wang, Cheng Lu, Yikai Wang, Fan Bao, Chongxuan Li, Hang Su, and Jun Zhu.
\newblock Prolificdreamer: High-fidelity and diverse text-to-3d generation with variational score distillation.
\newblock \emph{Advances in neural information processing systems}, 36:\penalty0 8406--8441, 2023.

\bibitem[Wu et~al.(2025)Wu, Lin, Zhang, Zeng, Yang, Bao, Qian, Zhu, Cao, Torr, et~al.]{wu2025direct3d}
Shuang Wu, Youtian Lin, Feihu Zhang, Yifei Zeng, Yikang Yang, Yajie Bao, Jiachen Qian, Siyu Zhu, Xun Cao, Philip Torr, et~al.
\newblock Direct3d-s2: Gigascale 3d generation made easy with spatial sparse attention.
\newblock \emph{arXiv preprint arXiv:2505.17412}, 2025.

\bibitem[Xiang et~al.(2025{\natexlab{a}})Xiang, Chen, Xu, Wang, Lv, Deng, Zhu, Dong, Zhao, Yuan, et~al.]{xiang2025native}
Jianfeng Xiang, Xiaoxue Chen, Sicheng Xu, Ruicheng Wang, Zelong Lv, Yu~Deng, Hongyuan Zhu, Yue Dong, Hao Zhao, Nicholas~Jing Yuan, et~al.
\newblock Native and compact structured latents for 3d generation.
\newblock \emph{arXiv preprint arXiv:2512.14692}, 2025{\natexlab{a}}.

\bibitem[Xiang et~al.(2025{\natexlab{b}})Xiang, Lv, Xu, Deng, Wang, Zhang, Chen, Tong, and Yang]{xiang2025structured}
Jianfeng Xiang, Zelong Lv, Sicheng Xu, Yu~Deng, Ruicheng Wang, Bowen Zhang, Dong Chen, Xin Tong, and Jiaolong Yang.
\newblock Structured 3d latents for scalable and versatile 3d generation.
\newblock In \emph{Proceedings of the Computer Vision and Pattern Recognition Conference}, pages 21469--21480, 2025{\natexlab{b}}.

\bibitem[Xu et~al.(2024)Xu, Cheng, Gao, Wang, Gao, and Shan]{xu2024instantmesh}
Jiale Xu, Weihao Cheng, Yiming Gao, Xintao Wang, Shenghua Gao, and Ying Shan.
\newblock Instantmesh: Efficient 3d mesh generation from a single image with sparse-view large reconstruction models.
\newblock \emph{arXiv preprint arXiv:2404.07191}, 2024.

\bibitem[Ye et~al.(2025)Ye, Wu, Lu, Chang, Guo, Zhou, Zhao, and Han]{ye2025hi3dgen}
Chongjie Ye, Yushuang Wu, Ziteng Lu, Jiahao Chang, Xiaoyang Guo, Jiaqing Zhou, Hao Zhao, and Xiaoguang Han.
\newblock Hi3dgen: High-fidelity 3d geometry generation from images via normal bridging.
\newblock \emph{arXiv preprint arXiv:2503.22236}, 3:\penalty0 2, 2025.

\bibitem[Zhang et~al.(2023)Zhang, Tang, Niessner, and Wonka]{zhang20233dshape2vecset}
Biao Zhang, Jiapeng Tang, Matthias Niessner, and Peter Wonka.
\newblock 3dshape2vecset: A 3d shape representation for neural fields and generative diffusion models.
\newblock \emph{ACM Transactions On Graphics (TOG)}, 42\penalty0 (4):\penalty0 1--16, 2023.

\bibitem[Zhang et~al.(2024)Zhang, Wang, Zhang, Qiu, Pang, Jiang, Yang, Xu, and Yu]{zhang2024clay}
Longwen Zhang, Ziyu Wang, Qixuan Zhang, Qiwei Qiu, Anqi Pang, Haoran Jiang, Wei Yang, Lan Xu, and Jingyi Yu.
\newblock Clay: A controllable large-scale generative model for creating high-quality 3d assets.
\newblock \emph{ACM Transactions on Graphics (TOG)}, 43\penalty0 (4):\penalty0 1--20, 2024.

\bibitem[Zhao et~al.(2025)Zhao, Lai, Lin, Zhao, Liu, Yang, Feng, Yang, Zhang, Yang, et~al.]{zhao2025hunyuan3d}
Zibo Zhao, Zeqiang Lai, Qingxiang Lin, Yunfei Zhao, Haolin Liu, Shuhui Yang, Yifei Feng, Mingxin Yang, Sheng Zhang, Xianghui Yang, et~al.
\newblock Hunyuan3d 2.0: Scaling diffusion models for high resolution textured 3d assets generation.
\newblock \emph{arXiv preprint arXiv:2501.12202}, 2025.

\end{thebibliography}

\end{document}